\documentclass[letterpaper, 10 pt, conference]{ieeeconf}  

\IEEEoverridecommandlockouts                              

\usepackage{makecell}
\usepackage{graphicx}
\usepackage{xcolor}
\usepackage{soul}
\usepackage[ruled,vlined]{algorithm2e}
\usepackage{amsmath}
\usepackage{gensymb}
\usepackage{bbm}
\usepackage{hyperref}
\title{\LARGE \bf
Reinforcement co-Learning of Deep and Spiking Neural Networks for Energy-Efficient Mapless Navigation with Neuromorphic Hardware}

\author{Guangzhi Tang, Neelesh Kumar, and Konstantinos P. Michmizos
\thanks{*This work is supported by Intel's NRC Grant Award}
\thanks{GT, NK and KM are with the Computational Brain Lab, Department of Computer Science, Rutgers University, New Jersey, USA.
        {\tt\small konstantinos.michmizos@cs.rutgers.edu}}%
}

\begin{document}

\maketitle
\thispagestyle{empty}
\pagestyle{empty}

\begin{abstract}
Energy-efficient mapless navigation is crucial for mobile robots as they explore unknown environments with limited on-board resources. Although the recent deep reinforcement learning (DRL) approaches have been successfully applied to navigation, their high energy consumption limits their use in several robotic applications. Here, we propose a neuromorphic approach that combines the energy-efficiency of spiking neural networks with the optimality of DRL and benchmark it in learning control policies for mapless navigation. Our hybrid framework, spiking deep deterministic policy gradient (SDDPG), consists of a spiking actor network (SAN) and a deep critic network, where the two networks were trained jointly using gradient descent. The co-learning enabled synergistic information exchange between the two networks, allowing them to overcome each other's limitations through a shared representation learning. To evaluate our approach, we deployed the trained SAN on Intel's Loihi neuromorphic processor. When validated on simulated and real-world complex environments, our method on Loihi consumed 75 times less energy per inference as compared to DDPG on Jetson TX2, and also exhibited a higher rate of successful navigation to the goal, which ranged from 1\% to 4.2\% and depended on the forward-propagation timestep size. These results reinforce our ongoing efforts to design brain-inspired algorithms for controlling autonomous robots with neuromorphic hardware. 
\end{abstract}
\section{Introduction}
The ability of a mobile robot to navigate autonomously becomes increasingly important with the complexity of the unknown environment that it explores. Traditionally, navigation has been relying on global knowledge in the form of maps of the environment \cite{meyer2003map}. Yet, for many applications in need of effective navigation, the construction of an informative map is prohibitively expensive, due to real-time requirements and the limited energy resources \cite{niroui2019deep,lahijanian2018resource}.
The recent introduction of deep reinforcement learning (DRL) methods, such as deep deterministic policy gradient (DDPG) \cite{lillicrap2015continuous}, enabled learning of optimal control policies for mapless navigation, where the agent navigates using its local sensory inputs and limited global knowledge \cite{tai2017virtual,choi2019deep,zhu2017target}. 
The optimality of DRL, however, comes at a high-energy cost. Given that the growing complexity of mobile robot applications is hard to be continuously offset by equivalent increases in on-board energy sources, there is an unmet need for low-power solutions to robotic mapless navigation.
\par 
Energy-efficiency is currently the main advantage demonstrated by spiking neural networks (SNN), an emerging brain-inspired alternative architecture to deep neural networks (DNN) in which neurons compute asynchronously and communicate through discrete events called spikes\cite{maass1997networks}. We and others have recently shown how the realization of SNNs in a neuromorphic processor results in low-power solutions for mobile robots, ranging from localization and mapping of mobile robots \cite{tang2019spiking} on Intel's Loihi \cite{davies2018loihi} to planning \cite{fischl2017path} and control \cite{blum2017neuromorphic}. For mapless navigation, most SNN-based approaches employ a reward modulated learning, where a global reward signal drives the local synaptic weight updates \cite{bing2018end,guan2020unsupervised}. Despite the biological plausibility of this learning rule, it suffers from catastrophic forgetting and a lack of policy evaluation \cite{bing2018end}, which limit the learning of policies in complex real-world environments.
\par 
\begin{figure}
\vspace{5.2pt}
\centering
\includegraphics[scale=1.0]{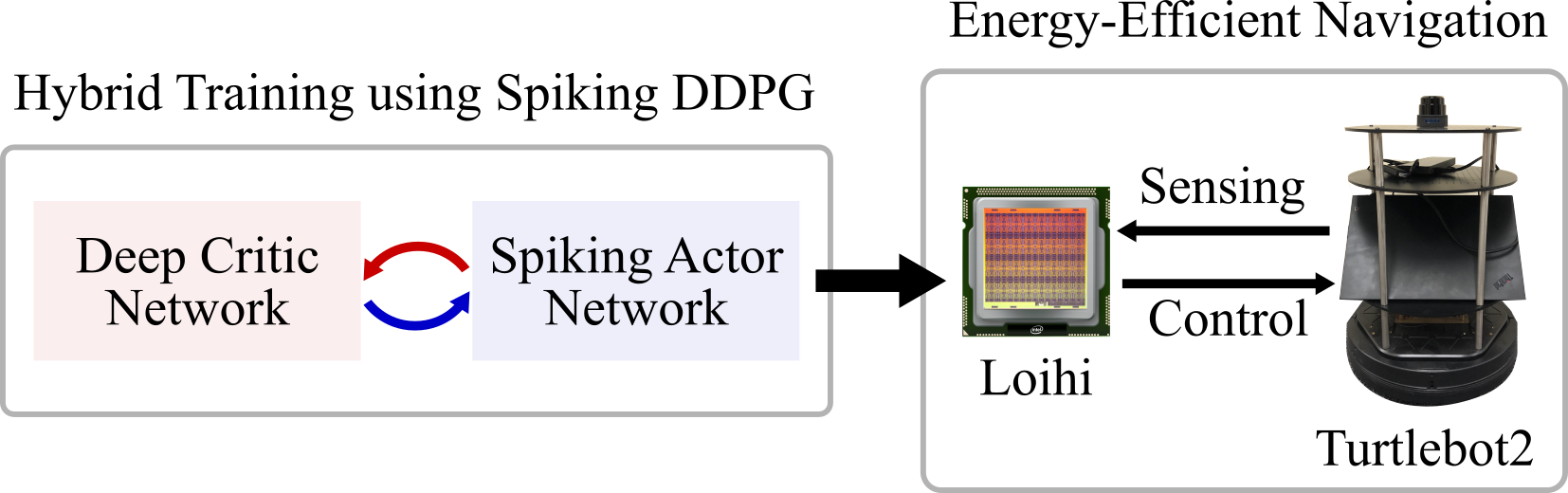}
\caption{Overview of the proposed hybrid framework, Spiking DDPG, that consisted of a spiking actor network and a deep critic network that were co-trained. The trained SAN was deployed on Intel's Loihi neuromorphic chip for controlling a mobile robot in simulated and real-world complex environments. The resulted mapless navigation compared favorably with the state-of-the-art methods, in both energy-efficiency and accuracy.}
\label{fig:Overview}
\end{figure}
Interestingly, DRL methods are well-versed in overcoming catastrophic forgetting through memory replay, and provide systematic evaluation of policies \cite{lillicrap2015continuous}. That led us to wonder if, and to what extent, could we combine the advantages of two emerging methodologies, namely the energy-efficiency of an SNN with the computational capabilities of a DNN. 
Recent efforts to combine these two architectures have relied on directly converting the trained DNNs to SNNs using techniques such as weight-scaling \cite{diehl2016conversion}. Such methods, though, require larger time-steps for inference \cite{patel2019improved}, which becomes particularly problematic in mobile robots that need to make decisions in real-time. To overcome this limitation, one possibility is to directly train SNNs using gradient-descent techniques such as spatiotemporal backpropagation (STBP) \cite{wu2018spatio}. This method has demonstrated faster inference while exhibiting state-of-the-art performance for a wide range of classification tasks \cite{shrestha2018slayer,wu2018spatio}. However, the limited ability of spiking neurons to represent high precision action-values would result in the prediction of the same action-values for different inputs and prevent the generation of corrective errors during training. Given the ability of the DRL approaches to represent high precision action-values, a fascinating possibility for developing a neuromorphic solution to mapless navigation in complex environments is to train the SNN in conjunction with a deep network.
 
\par 
In this paper, we propose Spiking DDPG (SDDPG), an energy-efficient neuromorphic method that uses a hybrid SNN/DNN framework to learn optimal policies\footnote{Code: https://github.com/combra-lab/spiking-ddpg-mapless-navigation}, and benchmark our approach in mapless robotic navigation in real-world environments (Fig. \ref{fig:Overview}). Like its deep network counterpart, SDDPG had separate networks to represent policy and action-value: a spiking actor network (SAN), to infer actions from the robot states, and a deep critic network, to evaluate the actor. The two networks in this architecture were trained jointly using gradient-descent. To train the SAN, we introduce an extension of STBP, which allowed us to faithfully deploy the trained SAN on Intel's Loihi. We evaluated our method through comparative analyses for performance and energy-efficiency with respect to DDPG in simulated and real-world complex environments. The SDDPG on Loihi consumed 75 times less energy per inference when compared against DDPG on Jetson TX2, while also achieving a higher rate of successfully navigating to the goal.

\section{Methods}
\subsection{Spiking Deep Deterministic Policy Gradient (SDDPG)}

We propose the SDDPG algorithm to learn the optimal control policies for mapping a given state of the robot $s = \{G_{dis},\ G_{dir},\ \nu,\ \omega,\ S\}$ to the robot action $a = \{\nu_L,\ \nu_R\}$, where $G_{dis}$ and $G_{dir}$ are the relative distance and direction from the robot to the goal; $\nu$ and $\omega$ are the linear and angular velocities of the robot; S is the distance observations from the laser range scanner; $\nu_L$ and $\nu_R$ are the left and right wheel speeds of the differential drive mobile robot. 

The hybrid framework consisted of a spiking actor network (SAN) and a deep critic network (Fig. \ref{fig:SAN}). During training, the SAN
generated an action $a$ for a given state $s$, which was then fed to the critic network for predicting the associated action-value $Q(s,a)$. The SAN was trained to predict the action for maximizing this Q value. The critic network, in turn, was trained to minimize the temporal difference (TD) error for action-value, as described in \cite{lillicrap2015continuous}. To update the action-value, we used a reward function adopted from \cite{tai2017virtual}:
 \begin{equation}
      \mathnormal{R} =
  \begin{cases}
                                   \mathnormal{R}_{goal} & \text{if $G_{dis} < G_{th}$} \\
                                    \mathnormal{R}_{obstacle} & \text{if $O_{dis} < O_{th}$} \\
  A*(G_{dis}(t) - G_{dis}(t-1)) & \text{otherwise}
  \end{cases}
  \end{equation}
  where $R_{goal}$ and $R_{obstacle}$ are the positive and negative rewards, respectively; $O_{dis}$ is the distance to the obstacle; $A$ is an amplification factor; $G_{th}, \ O_{th}$ are the thresholds. The reward function encourages the robot to move towards the goal during exploration, which facilitates training. 

For inference, we deployed the trained SAN on Loihi (see \textit{II.D}), to predict the action to navigate the robot to the goal. We give the mathematical formalism for the inference and training phases in the next two sections.

\subsection{Spiking Actor Network (SAN)}
The building block of the SAN was the leaky-integrate-and-fire (LIF) model of a spiking neuron. Specifically, we updated the states of the $i^{th}$ neuron at timestep $t$ in two stages. First, we integrated the input spikes into synaptic current as follows,
\begin{equation}
    c_i(t) = d_c\cdot c_i(t-1) + \Sigma_jw_{ij}o_j(t) \label{eq:1}
\end{equation}
where $c$ is the synaptic current, $d_c$ is the decay factor for the current, $w_{ij}$ is the weight of the connection from the $j^{th}$ presynaptic neuron and $o_j$ is a binary variable (0 or 1) which indicates the spike event of the $j^{th}$ presynaptic neuron.
\par
Second, we integrated the synaptic current into the membrane voltage of the neuron as per equation \eqref{eq:2}. Subsequently, the neuron fired a spike if the membrane voltage exceeded the threshold.
\begin{equation}
\begin{gathered}
v_i(t) = d_v\cdot v_i(t-1) + c_i(t), \quad \textrm{if } v_i(t-1) < V_{th} \\
 o_i(t) = 1\textrm{ \&  } v_i(t) = 0, \quad \textrm{otherwise}
    \label{eq:2}
\end{gathered}
\end{equation}
where $v$ is the membrane voltage, $d_v$ is the decay factor for the voltage and $V_{th}$ is the firing threshold. 
\par

\begin{figure}
\vspace{5.2pt}
\centering
\includegraphics[scale=1.0]{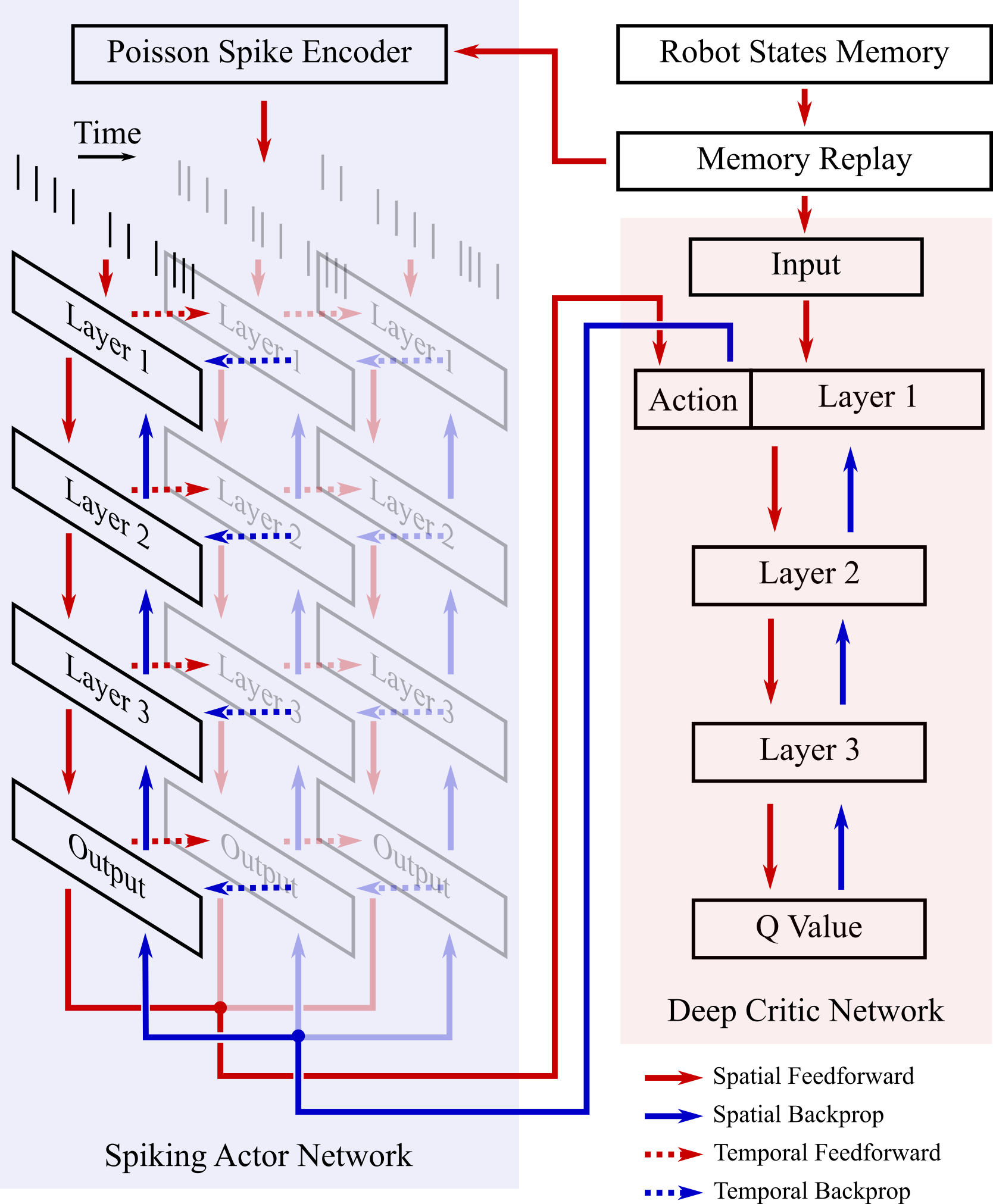}
\caption{Hybrid training of multilayered SAN and Deep Critic Network.}
\label{fig:SAN}
\end{figure}

The LIF neurons formed a fully connected multilayered SAN (Fig. \ref{fig:SAN}). The network was driven by discrete Poisson spikes that encoded the continuous state variables. This was done by generating a spike at each timestep with probability proportional to the value of the state variables at that time.  After $T$ timesteps, we decoded the rescaled average spike count of the output layer neurons (\textbf{Action}) to left and right wheel speeds of the robot (Algorithm \ref{alg:forward}). 

\begin{algorithm}[h]
\SetAlgoLined
\textbf{Output: } Left and right wheel speeds $\nu_L$, $\nu_R$\

\textbf{Require:} Maximum timestep, $T$; Network depth, $l$\

\textbf{Require:} Min and max wheel speeds $\nu_{min}$, $\nu_{max}$\

\textbf{Require:} $\textbf{W}^{(i)}$, $i\in\{1,...,l\}$, the weight matrices\

\textbf{Require:} $\textbf{b}^{(i)}$, $i\in\{1,...,l\}$, the bias parameters\

\textbf{Require:} $\textbf{X}^{(i)}$, $i\in\{1,...,T\}$, the input spike trains

\For{t=1,...,T}{
$\textbf{o}^{(t)(0)}=\textbf{X}^{(t)}$\;
    \For{k=1,...,l}{
    $\textbf{c}^{(t)(k)} = d_c \cdot \textbf{c}^{(t-1)(k)}+ \textbf{W}^{(k)}\textbf{o}^{(t)(k-1)} + \textbf{b}^{(k)}$\;
    $\textbf{v}^{(t)(k)} = d_v \cdot \textbf{v}^{(t-1)(k)}\cdot (1-\textbf{o}^{(t-1)(k)}) + \textbf{c}^{(t)(k)}$\;
    $\textbf{o}^{(t)(k)} = Threshold(\textbf{v}^{(t)(k)})$\;
    }
    $\textbf{SpikeCount}^{(t)} = \textbf{SpikeCount}^{(t-1)} + \textbf{o}^{(t)(l)}$\;
}
$\textbf{Action} = \textbf{SpikeCount}^{(T)} /\ T$\\
$\nu_L = \textbf{Action}[0] * (\nu_{max} - \nu_{min}) + \nu_{min}$\\
$\nu_R = \textbf{Action}[1] * (\nu_{max} - \nu_{min}) + \nu_{min}$
 \caption{Forward propagation through SAN}
 \label{alg:forward}
\end{algorithm}

\subsection{Direct Training of SAN with Back-propagation}

We extended the STBP to directly train our SAN for learning the optimal policy. The original STBP is limited to training networks containing simplified LIF neurons that have only one state variable (voltage). Here, we extended it to LIF neurons with two internal state variables (current and voltage), defined in the equations \eqref{eq:1} and \eqref{eq:2}. This was done so that we could deploy our trained model on Loihi, which implements such a two-state neuron model.

Since the threshold function that defines a spike is non-differentiable, the STBP algorithm requires a pseudo-gradient function to approximate the gradient of a spike. We chose the rectangular function (defined in equation \eqref{eq:pseudo}) as our pseudo-gradient function since it demonstrated the best empirical performance in \cite{wu2018spatio}. 
\begin{equation}
      \mathnormal{z(v)} =
  \begin{cases}
  \mathnormal{a}_{1} & \text{if $|v - V_{th}| < a_2$} \\
  0 & \text{otherwise}
  \end{cases}
   \label{eq:pseudo}
  \end{equation}
 where $z$ is the pseudo-gradient, $a_1$ is the amplifier of the gradient, $a_2$ is the threshold window for passing the gradient.

At the end of the forward propagation of the SAN, the computed \textbf{Action} was fed to the $n^{th}$ layer of the critic network,  which in turn generated the predicted $Q$ value. The SAN was trained to predict the action for which the trained critic network generated the maximum $Q$ value. To do so, we trained our SAN to minimize the loss $L$ defined in equation \eqref{eq:loss}, using gradient descent.
\begin{equation}
    L = -Q. \label{eq:loss}
\end{equation}
The gradient on the $n^{th}$ layer of the critic network was:
\begin{equation}
    \nabla _{\textbf{Action}} L = \textbf{W}_c^{(n+1)'}\cdot\nabla_{a^{(n+1)}}L, \label{eq:3}
\end{equation}
where $a^{n+1}$ is the output of the $(n+1)^{th}$ layer of the critic network before being passed through the non-linear activation function, such as $ReLU$, and $\textbf{W}_c^{(n+1)}$ are the critic network weights at the $(n+1)^{th}$ layer. 

This gradient was then backpropagated to the SAN. To describe the complete gradient descent, we separate our analysis into two cases: i) the last forward propagation timestep, $t = T$, and ii) all the previous timesteps, $t < T$.  \\ 
\textbf{Case 1}: for $t = T$.
\\
At the output layer $l$, we have:
\begin{equation}
    \begin{gathered}
    \nabla _{\textbf{SpikeCount}^{(t)}} L = \frac{1}{T}\cdot\nabla _{\textbf{Action}} L\\
    \nabla _{\textbf{o}^{(t)(k)}} L = \nabla _{\textbf{SpikeCount}^{(t)}} L \label{eq:4}
    \end{gathered}
\end{equation}
Then for each layer, $k = l$ down to $1$: \\
\begin{equation}
    \begin{gathered}
    \nabla _{\textbf{v}^{(t)(k)}} L = z(\textbf{v}^{(t)(k)})\cdot\nabla _{\textbf{o}^{(t)(k)}} L \\
    \nabla _{\textbf{c}^{(t)(k)}} L = \nabla _{\textbf{v}^{(t)(k)}} L \\
      \nabla _{\textbf{o}^{(t)(k-1)}} L = \textbf{W}^{(k)'}\cdot \nabla _{\textbf{c}^{(t)(k)}} L \label{eq:5}
    \end{gathered}
\end{equation}
\\
\textbf{Case 2}: for $t < T$.
\\
At the output layer $l$, we have:
\begin{equation}
    \begin{gathered}
    \nabla _{\textbf{SpikeCount}^{(t)}} L = \nabla _{\textbf{SpikeCount}^{(t+1)}} L = \frac{1}{T}\cdot\nabla _{\textbf{Action}} L \\
     \nabla _{\textbf{o}^{(t)(l)}} L = \nabla _{\textbf{SpikeCount}^{(t)}} L \label{eq:6}
    \end{gathered}
\end{equation}
Then for each layer, $k = l$ down to $1$: 
\begin{equation}
\begin{gathered}
\begin{aligned}
    \nabla _{\textbf{v}^{(t)(k)}} L = z(\textbf{v}^{(t)(k)})\cdot\nabla _{\textbf{o}^{(t)(k)}} L + \\
    d_v(1-\textbf{o}^{(t)(k)})\cdot\nabla _{\textbf{v}^{(t+1)(k)}} L 
\end{aligned}
\\
    \nabla _{\textbf{c}^{(t)(k)}} L = \nabla _{\textbf{v}^{(t+1)(k)}} L + 
    d_c\nabla _{\textbf{c}^{(t+1)(k)}} L  \\
    \nabla _{\textbf{o}^{(t)(k-1)}} L = \textbf{W}^{(k)'}\cdot \nabla _{\textbf{c}^{(t)(k)}} L \label{eq:7}
\end{gathered}
\end{equation}
In this case, the gradients with respect to voltage and current had additional terms as compared to case 1, reflecting the temporal gradients backpropagated from the future timesteps.

By collecting the gradients backpropagated from all the timesteps (computed in the above two cases), we can compute the gradient of the loss with respect to the network parameters, $\nabla _{\textbf{W}^{(k)}} L$, $\nabla _{\textbf{b}^{(k)}} L$ for each layer $k$, as below:

\begin{equation}
    \begin{gathered}
    \nabla _{\textbf{W}^{(k)}} L = \sum_{t=1}^{T}\textbf{o}^{(t)(k-1)}\cdot\nabla _{\textbf{c}^{(t)(k)}} L \\
    \nabla _{\textbf{b}^{(k)}} L = \sum_{t=1}^{T}\nabla _{\textbf{c}^{(t)(k)}} L \label{eq:9}
    \end{gathered}
\end{equation}
We updated the network parameters every $T$ timesteps.

\begin{figure}
\vspace{5.2pt}
\centering
\includegraphics[scale=1.0]{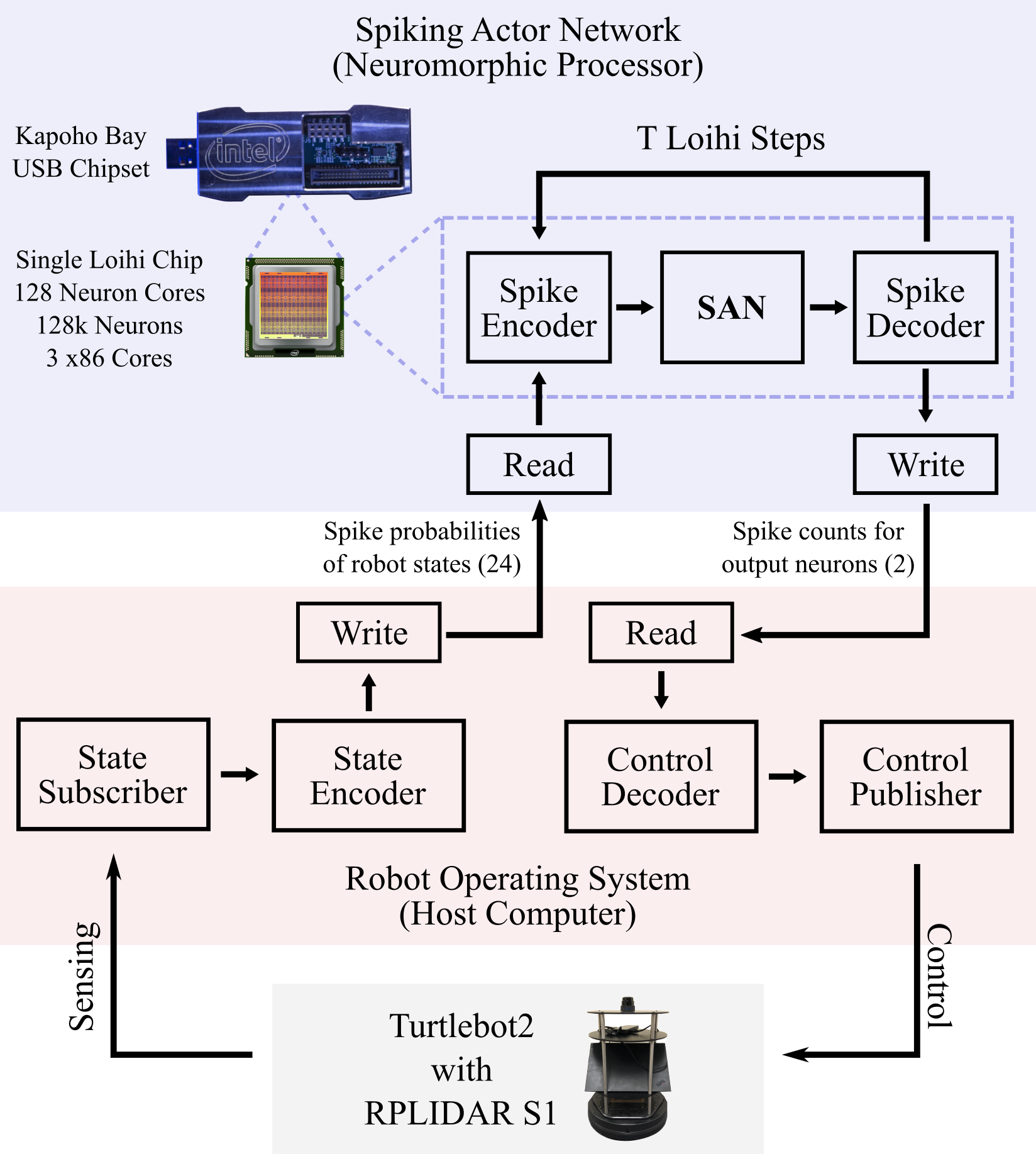}
\caption{Interaction between Loihi and the mobile robot through the robot operating system (ROS), for a single SNN inference. }
\label{fig:Loihi}
\end{figure}

\begin{figure}[b!]
\centering
\includegraphics[scale=1.0]{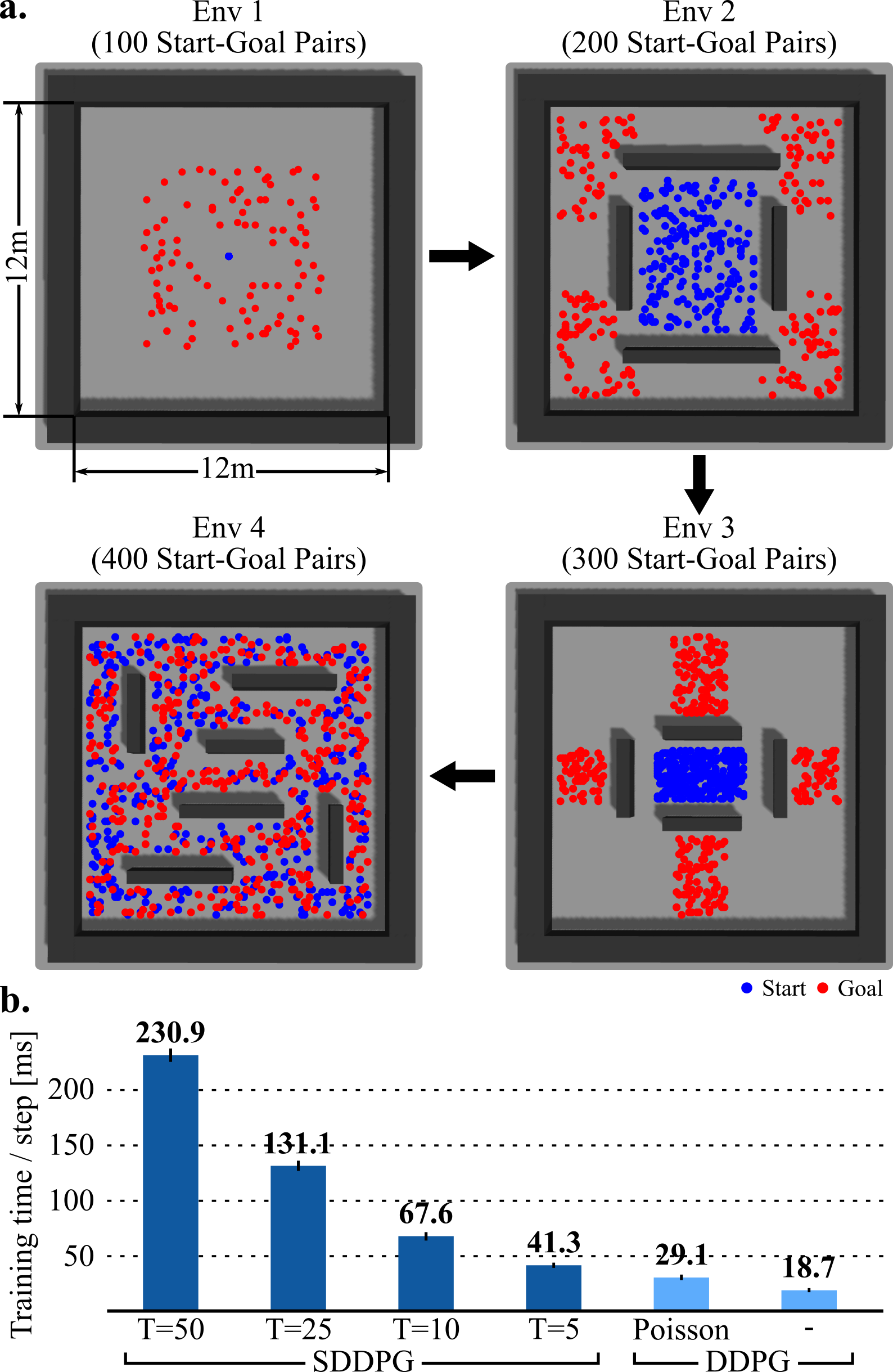}
\caption{\textbf{a.} The four training environments of increasing complexity, with randomly generated start and goal locations enabled curriculum learning. \textbf{b.} Training time per execution step of SDDPG decreased with decreasing T and approached the training time for DDPG. }
\label{fig:Training}
\end{figure}

\subsection{SAN Realization on Loihi Neuromorphic Processor}
We realized our trained SAN on Intel's Loihi. Since Loihi supports 8 bits integer weights, we introduce a layer-wise rescaling technique for mapping the trained SNNs with higher weight precisions onto the chip. We rescaled the weights and voltage threshold of each layer using equation \eqref{eq:loihi}, while maintaining their spike outputs by fixing the weight-threshold ratio. As a benefit of training our network with the neuron model that Loihi supports, all other hyperparameters remained the same as the ones used in training. 
\begin{equation}
    \begin{gathered}
    r^{(k)} = \frac{W_{max}^{(k)(loihi)}}{W_{max}^{(k)}} \\
  \textbf{W}^{(k)(loihi)} = round(r^{(k)} \cdot \textbf{W}^{(k)}) \\
  V_{th}^{(k)(loihi)} = round(r^{(k)} \cdot V_{th}) \label{eq:loihi}
    \end{gathered}
\end{equation}
where $r^{(k)}$ is the rescale ratio of layer $k$, $W_{max}^{(k)(loihi)}$ is the maximum weight that Loihi supports, $W_{max}^{(k)}$ is the maximum weight of layer $k$ of the trained network, $\textbf{W}^{(k)(loihi)}$ are the rescaled weights on Loihi, and $V_{th}^{(k)(loihi)}$ is the rescaled voltage threshold on Loihi.

We also introduce an interaction framework for Loihi to control the mobile robot in real-time through the robot operating system (ROS) (Fig. \ref{fig:Loihi}). We encoded the robotic states obtained from ROS into Poisson spikes, and decoded the output spikes for robot control. The encoding and decoding modules were deployed on the low-frequency x86 cores that Loihi has for interfacing with the on-chip networks during runtime. This avoided the need for communicating between Loihi and ROS directly through spikes, which reduced the data transfer load between Loihi and ROS.  
\begin{figure*}[t]
\vspace{5.2pt}
\centering
\includegraphics[scale=1.0]{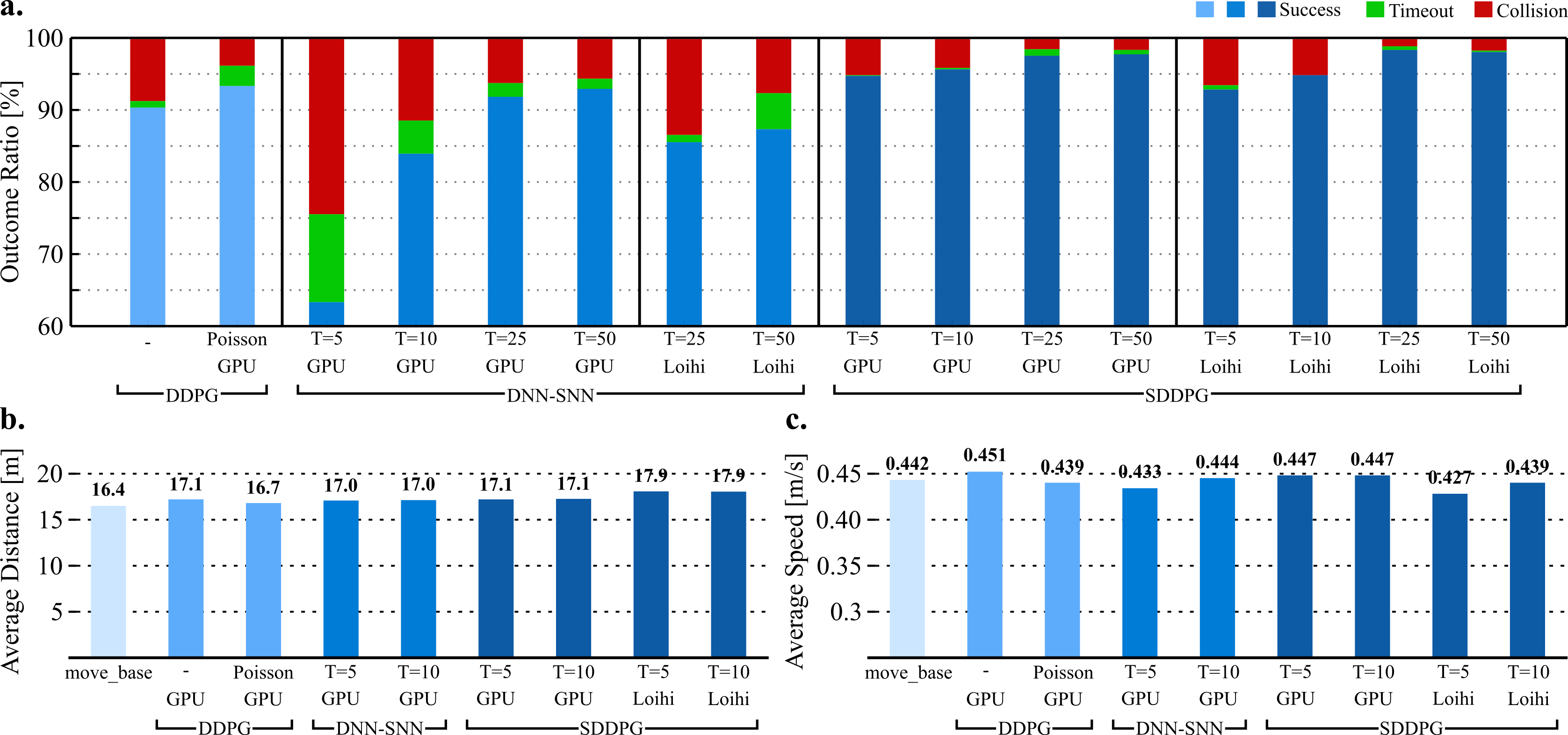}
\caption{Comparing SDDPG with other navigation methods in a complex test environment in the simulator, with 200 randomly generated start and goal positions. \textbf{a.} SDDPG has a higher rate of successfully navigating to the goal than the DNN-SNN and DDPG methods. There was no decrease in its performance when deployed on Loihi. \textbf{b}, \textbf{c.} SDDPG results in similar route quality (average distance and speed) as the other navigation methods. Results are averaged over five trained models for each method.} 
\label{fig:Eval1}
\end{figure*}

\section{Experiments and Results}
\subsection{Experimental Setup}
We trained and validated our method on Turtlebot2 platform equipped with an RPLIDAR S1 laser range scanner (range: 0.2-40 m). The robot's field of view was set to front-facing 180 degrees with 18 range measurements, each with 10 degrees of resolution. Training was performed in the Gazebo simulator and validation was done in both the simulation and real-world environments. We used the ROS as a middleware for both the training and validation. The neuromorphic realization was performed on Intel's Kapoho-Bay USB chipset containing two Loihi chips.

\subsection{Training in Simulator}
During training, the agent sequentially navigated 4 environments of increasing complexities (Fig. \ref{fig:Training}a). The start and goal locations were sampled randomly from specific places in the 4 environments.  The increase in difficulty across the 4 environments was due to the added obstacles and the different start-goal pairs. This encouraged the robot to build upon previously learned simple policies in easier environments (Env 1 and 2) and gradually learn complex policies for navigating in difficult environments (Env 3 and 4). This form of curriculum training has been shown to result in better generalization and faster convergence \cite{florensa2017reverse,bengio2009curriculum}.

\begin{table}
    \caption{Hyperparameters for training SDDPG}
    \centering
    \begin{tabular}{l c}
    \hline
    Parameters & Values\\
    \hline
    Neuron parameters ($V_{th},d_{c},d_{v}$) & 0.5, 0.5, 0.75\\
    Pseudo-gradient function parameters ($a_1,a_2$) & 1.0, 0.5\\
    Neurons per hidden layer for SAN and critic net & 256, 512\\
    Batchsize & 256\\
    Learning rate for training SAN and critic net  & $10^{-5}$, $10^{-4}$\\
    Goal and collision reward ($R_{goal},R_{obstacle}$) & 30, -20\\
    Reward amplification factor ($A$) & 15\\
    Reward thresholds ($G_{th}, O_{th}$) & 0.5, 0.35 m\\
    Wheel speed ($\nu_{min}, \nu_{max}$) & 0.05, 0.5 m/s\\
    \hline
    \end{tabular}
    \label{tab:hyper}
\end{table}

Each episode in training resulted in 3 possible outcomes: i) success: robot successfully navigated to the goal; ii) collision: robot collided with an obstacle; iii) timeout: navigation exceeded 1000 execution steps, with each execution step being 0.1s. Training was performed for a total of 200,000 execution steps across all 4 environments. To average out the effect of exploration during training and for fair comparison with the baselines, we trained 5 models corresponding to 5 sets of randomly initialized start and goal pairs. Moreover, to investigate the effect of the forward propagation time, $T$, we trained SDDPG corresponding to 4 different values of $T = {5, 10, 25, 50}$. Training time decreased with decreasing $T$ (Fig.  \ref{fig:Training}b), which partially overcame the limitation of high training time commonly associated with SNNs. The hyperparameters used for training are shown in Table \ref{tab:hyper}.

\subsection{Baselines for comparison}
We compared SDDPG with the following approaches:
\subsubsection{Map-based Navigation}
We used the widely used ROS navigation package move\_base consisting of a DWA (dynamic window approach) \cite{fox1997dynamic} local planner and a global planner based on Djikstra's algorithm. The map required for move\_base was constructed using GMapping \cite{grisetti2007improved}. The robot's maximum speed was set to 0.5 m/s, same as SDDPG. 

\subsubsection{DDPG}
The DDPG had the same network architecture as our SDDPG, with the SAN replaced by a deep actor network. To investigate the role of randomly generated Poisson spike inputs, we also compared our SDDPG against DDPG receiving inputs injected with Poisson noise (DDPG Poisson). To do this, we encoded the state inputs to Poisson spikes and then decoded it back to continuous state inputs. The baseline (DDPG/DDPG Poisson) and SDDPG methods were trained using the same hyperparameters. Training times for DDPG and SDDPG methods are shown in Fig. \ref{fig:Training}b.

\subsubsection{DNN to SNN Conversion (DNN-SNN)}
We converted a deep actor network trained using DDPG with Poisson noise, to an SNN, with same $T$ values as the SDDPG, using weight rescaling \cite{diehl2016conversion}. We determined the optimum rescale factor by computing the layer's maximum output over training duration, and then performing grid search around it \cite{patel2019improved}.

\begin{figure*}
\vspace{5.2pt}
\centering
\includegraphics[scale=1.0]{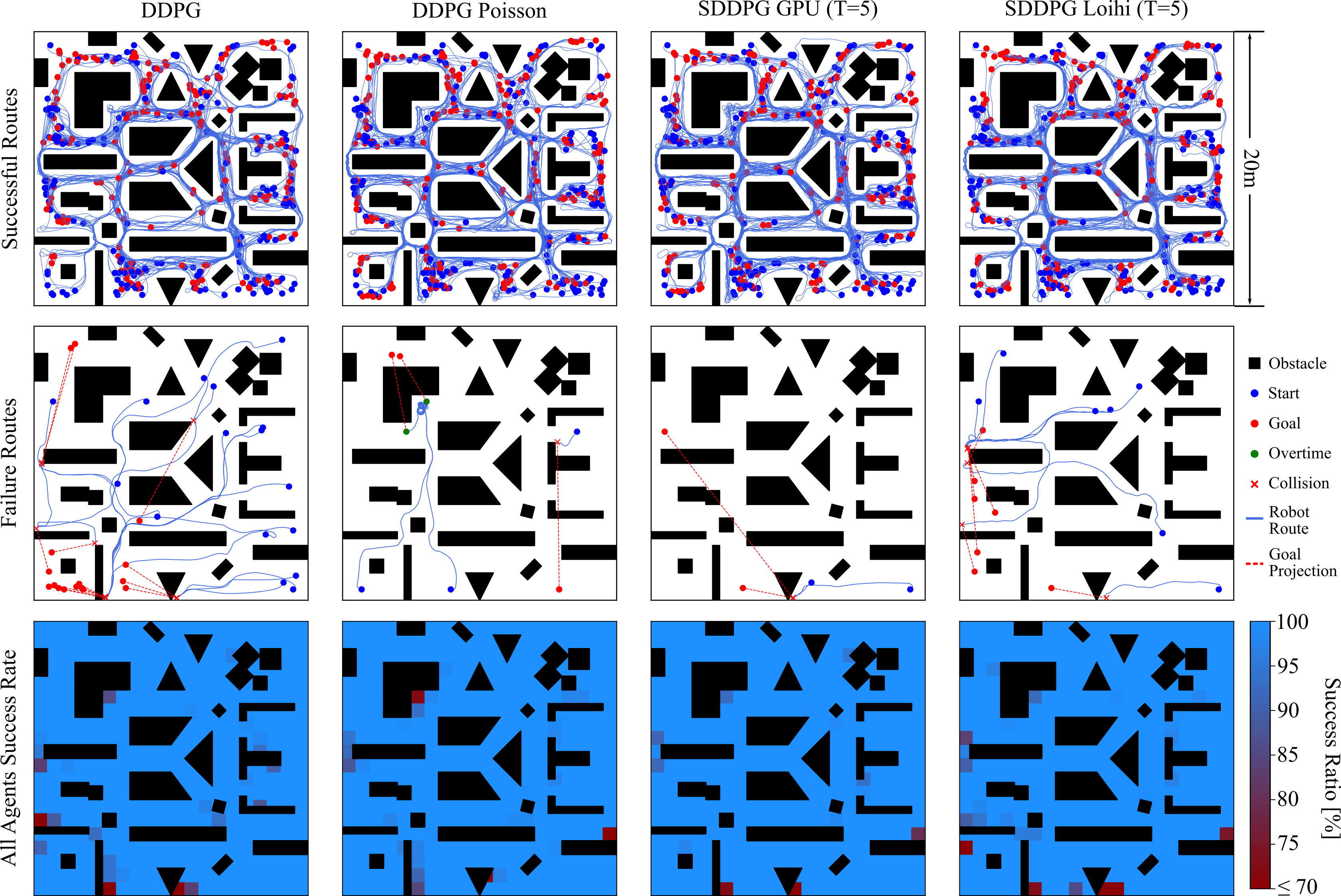}
\caption{Analyzing success and failure trajectories over 200 randomly sampled start and goal locations for the models with the highest success rate per method (upper and middle row). SDDPG fails at fewer locations than DDPG. The superior SDDPG performance is demonstrated in the heatmap showing the rate of successfully crossing each $1m \times 1m$ location in the test environment, computed across 5 models per method (bottom row).} 
\label{fig:Route}
\end{figure*}

\subsection{Evaluation in Simulator}
We evaluated our method in the Gazebo simulator in a $20m\times20m$ test environment (Fig. \ref{fig:Route}). To test the generalization capability of our method, we designed the test environment to be sufficiently different from the training environments in the following aspects: i) differently shaped obstacles (triangular, L-shaped); ii) narrower traversal passages (min. 0.75m for test, 1.75m for training);  iii) more densely organized obstacles. For an exhaustive evaluation, we generated 200 start and goal locations, sampled uniformly at random from all parts of the test environment with a minimum distance of 6m.  

We used the same start and goal locations to evaluate our method and all the baselines. We first compared the methods based on the rate of the three possible outcomes--success, collision, and timeout (Fig. \ref{fig:Eval1}a). Our method outperformed the DNN-SNN conversion method for all values of $T$, with the performance being substantially better for smaller values of $T$. Our method performed slightly better than DDPG, even when deployed on Loihi with low precision weights. To further inspect SDDPG's ability to navigate effectively, we compared its route quality against move\_base, as well as all the other navigation methods(Fig. \ref{fig:Eval1}b,c). Specifically, we computed the average distance and speed corresponding to the successful routes taken by each method. For fairness, we only considered the successful routes that had common start and goal positions across all methods. Despite not having access to the map, SDDPG achieved the same level of performance as the map-based method, move\_base. 

We then analyzed the trajectories of the routes that resulted in failures (collision or timeout) (Fig.  \ref{fig:Route}). The methods failed at the locations where it required the agents to move around the obstacles in the ways that it had never experienced in training.  To further investigate the failure locations, we generated a heatmap of the environment where the intensity of pixel corresponding to each $1m\times1m$ location was equal to the percentage of times the agent successfully crossed that location. The heatmap reveals that the SDDPG method failed at fewer locations than DDPG.

Although we did not explicitly target performance improvement over state-of-the-art, our results indicate that SDDPG had a slightly higher rate of successful navigation. A possible explanation is that the noise introduced by Poisson spike encoding of the state inputs helps the SDDPG networks in escaping the 'bad' local minima, in alignment with the results of \cite{plappert2017parameter}. The fact that DDPG Poisson performed better than DDPG further supports this reasoning. 
\begin{figure*}
\vspace{5.2pt}
\centering
\includegraphics[scale=1.0]{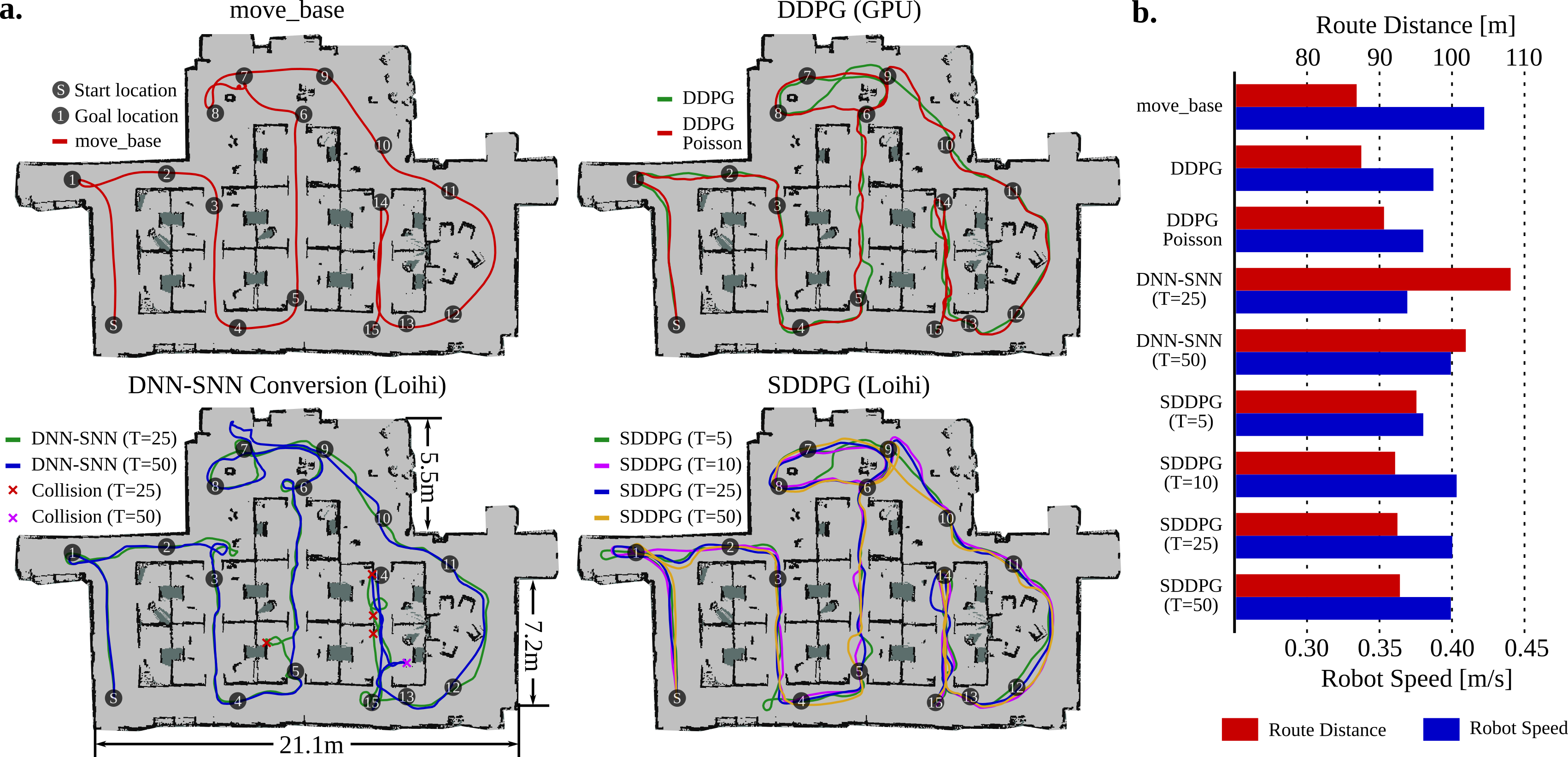}
\caption{Comparing SDDPG against other navigation methods in a complex real-world environment containing 15 sequential goal locations covering a large area of the environment. \textbf{a. }SDDPG successfully navigated to all the goal locations and took a similar route as the map-based navigation method (move\_base) and the other baseline methods. \textbf{b. } Route distance and speed metrics are similar for all tested methods.}
\label{fig:CBIM}
\end{figure*}
\subsection{Evaluation in Real-world}
We evaluated the navigation methods in a real-world environment to test the generalization capability of SDDPG (Fig.  \ref{fig:CBIM}a). The environment was an office setting consisting of cubicles and commonplace items such as chairs, desks, sofas, dustbins, and bookshelves. The space spanned over an area of approximately $215 m^2$, with the shortest passage being $0.9m$ in length. The robot was required to navigate to 15 goal locations placed sequentially to cover all the areas of the environment. We estimated the pose of the robot using $amcl$ \cite{fox1999monte}, based on the map generated by GMapping. Mapless navigation methods (DDPG, SDDPG) did not have access to this map. While the DNN-SNN method experienced several collisions in its route, the SDDPG method successfully navigated to all the goal locations and took a similar route as the map-based method (Fig.  \ref{fig:CBIM}a). Interestingly, SDDPG (T=10, 25, 50) exhibited slightly smoother movements than DDPG and DDPG Poisson (speed comparison in Fig.  \ref{fig:CBIM}b).
\begin{table}
    \caption{Power Performance across Hardware}
    \centering
    \begin{tabular}{c c c c c c}
    \hline
    Method & Device & Idle (W) & Dyn (W) & Inf/s & $\mu$J/Inf\\
    \hline
    DDPG & CPU & 13.400 & 58.802 & 6598 & 8910.84\\
    DDPG & GPU & 24.154 & 46.570 & 3053 & 15252.91\\
    DDPG & TX2(N) & 1.368 & 1.934 & 868 & 2227.41\\
    DDPG & TX2(Q) & 1.352 & 0.457 & 390 & 1171.70\\
    SDDPG(50) & Loihi & 1.087 & 0.017 & 125 & 131.99\\
    SDDPG(25) & Loihi & 1.087 & 0.014 & 203 & 67.29\\
    SDDPG(10) & Loihi & 1.087 & 0.011 & 396 & 28.47\\
    SDDPG(5) & Loihi & 1.087 & 0.007 & 453 & \textbf{15.53}\\
    \hline
    \end{tabular}
    \label{tab:hardware}
\end{table}
\subsection{Power Performance Measurement}
We performed a comparative analysis (Table \ref{tab:hardware}) of inference speed and energy consumption for the following mapless navigation solutions: i) DDPG  on E5-1660 CPU, ii) DDPG on Tesla K40 GPU, iii) DDPG on Jetson TX2, and iv) SDDPG on Loihi. We measured the average power consumed and the speed of performing inference for the robot states recorded during testing. We used tools that probed the on-board sensors to measure the power for each device: powerstat for CPU, nvidia-smi for GPU, sysfs for TX2, and energy probe for Loihi. The measurements for TX2 were taken for two of its power modes- the energy-efficient mode MAXQ (Q) and the high-performance mode MAXN (N). The energy cost per inference was obtained by dividing the power consumed over 1 second with the number of inferences performed in that second. Compared to DDPG running on the energy-efficient chip for DNNs, TX2 (Q),  SDDPG ($T=5$) was 75 times more energy-efficient while also having higher inference speed. There was a performance-cost tradeoff associated with SNNs of different timesteps $T$, suggesting that, for further improvement in navigation performance, SDDPG with larger $T$ may be preferred, albeit with a higher energy cost.

\section{Discussion and Conclusion}
In this paper, we propose a neuromorphic framework that combines the low power consumption and high robustness capabilities of SNNs with the representation learning capability of DNNs, and benchmark it in mapless navigation. While recent efforts on integrating the two architectures have focused on training the two networks separately \cite{patel2019improved,pei2019towards}, we present here a method to train them in conjunction with each other. Our training approach enabled synergistic information exchange between the two networks, allowing them to overcome each other's limitations through shared representation learning; This resulted in an optimal and energy-efficient solution to mapless navigation when deployed on a neuromorphic processor. Such efforts can complement the neuromorphic hardware that currently allow joint inference such as the Tianjic chip \cite{pei2019towards}, and spur the development of hybrid neuromorphic chips for energy-efficient joint training.

Our method was 75 times more energy-efficient than DDPG running on a low-power edge device for DNNs (TX2). This superior performance comes not only from the asynchronous and event-based computations provided by the SNNs, but also from the ability of our method to train the SNNs for lower values of $T$ with very little loss in performance. This is, however, not the case with the DNN-SNN conversion method, which required 5 times more timesteps and 4.5 times more energy to reach the same level of performance as SDDPG ($T$=5). This gain in energy-efficiency could enable our method to effectively navigate mobile service robots with limited on-board resources in domestic, industrial, medical, or disaster-relief applications. Further decreases in energy cost may be achieved by coupling our method with low-cost mapless localization methods, such as the ones based on active beacons \cite{yun2006robust}, and by utilizing analog memristive neuromorphic processors \cite{ankit2017resparc} that are orders of magnitudes more efficient than their digital counterparts.   
While most demonstrations of the SNN advantages focus on the energy gains and come at a cost of a drop in performance \cite{tang2019spiking,fischl2017path,blum2017neuromorphic}, this is perhaps the first time that an energy-efficient method also demonstrates better accuracy than the current state of the art, at least in the tested robotic navigation tasks. The accuracy increase is partly due to our hybrid training approach that helped to overcome the SNN limitation in representing high precision values, which led to better optimization. Moreover, the SNN's inherently noisy representation of its inputs in the spatiotemporal domain might have also contributed to escaping 'bad' local minima. These superior SDDPG results suggest reinforcement learning as a training paradigm where the energy-efficient SNNs may also realize their promises for computational robustness and versatility.     

Overall, this work supports our ongoing efforts to develop solutions for real-time energy-efficient robot navigation. Our mapless solution can complement the current map-based approaches for generating more reliable control policies in applications where maps can be easily acquired. In addition, our general hybrid framework can be used to solve a variety of tasks, paving the way for fully autonomous mobile robots. 
\bibliographystyle{IEEEtran}
\bibliography{myref}
\end{document}